\documentclass[letterpaper, 10 pt, conference]{ieeeconf}  

\IEEEoverridecommandlockouts                              

\usepackage{graphicx} 
\usepackage[dvipsnames]{xcolor}
\usepackage{biblatex}
\usepackage{hyperref}
\usepackage{adjustbox}
\usepackage{amsmath}
\usepackage[ruled,vlined]{algorithm2e}
\usepackage[font=footnotesize]{caption}
\usepackage{subcaption}
\usepackage{float}
\usepackage[baseline]{euflag}

\title{\LARGE \bf
AgriSORT: A Simple Online Real-time Tracking-by-Detection framework for robotics in precision agriculture
}

\author{Leonardo Saraceni, Ionut M. Motoi, Daniele Nardi, Thomas A. Ciarfuglia
\thanks{\euflag \quad This work has been partially supported by the European Commission under the grant agreement number 101016906 – Project CANOPIES}
\thanks{\euflag \quad This work has been partially supported by project AGRITECH Spoke 9
- Codice progetto MUR: AGRITECH ”National Research Centre for Agricultural
Technologies” - CUP CN00000022, of the National Recovery and Resilience Plan
(PNRR) financed by the European Union ”Next Generation EU”.}%
\thanks{This work has been partially supported by Sapienza University of Rome as part of the work for project \textit{H\&M: Hyperspectral and Multispectral Fruit Sugar Content Estimation for Robot Harvesting Operations in Difficult Environments}, Del. SA n.36/2022.}%
}

\newcommand{\pageheader}{\footnotesize \textbf{This work has been submitted to the IEEE for possible publication. Copyright may be transferred without notice, after which this version may no longer be accessible.}}

\AtBeginShipout{
  \AtBeginShipoutUpperLeft{
    \put(51,-35){
      \parbox{\textwidth}{\pageheader\hfill}
    }
  }
}

\begin{document}

\maketitle
\thispagestyle{empty}
\pagestyle{empty}

\begin{abstract}

The problem of multi-object tracking (MOT) consists in detecting and tracking all the objects in a video sequence while keeping a unique identifier for each object. It is a challenging and fundamental problem for robotics. In precision agriculture the challenge of achieving a satisfactory solution is amplified by extreme camera motion, sudden illumination changes, and strong occlusions. Most modern trackers rely on the appearance of objects rather than motion for association, which can be ineffective when most targets are static objects with the same appearance, as in the agricultural case. To this end, on the trail of SORT \cite{Bewley_2016}, we propose AgriSORT, a simple, online, real-time tracking-by-detection pipeline for precision agriculture based only on motion information that allows for accurate and fast propagation of tracks between frames. The main focuses of AgriSORT are efficiency, flexibility, minimal dependencies, and ease of deployment on robotic platforms. We test the proposed pipeline on a novel MOT benchmark specifically tailored for the agricultural context, based on video sequences taken in a table grape vineyard, particularly challenging due to strong self-similarity and density of the instances. Both the code and the dataset are available for future comparisons.

\end{abstract}

\section{Introduction}
\label{sec::intro}
The deployment of robots in the precision agriculture context (Fig. \ref{fig::robot}) has seen rapid growth because of their potential to reduce the high cost of repetitive labor and wasted resources. Some possible tasks are weeding, fruit detection and harvesting, yield estimation, and fertilizer or pesticide application. Many works \cite{jin2022novel, saleem2021automation} in literature have addressed the problem of detecting crops in single frames using various sensors. Although analyzing individual frames can offer insight into a crop, this is not enough for many tasks. In fact, since changing point of view may lead to an entirely different measurement, algorithms need to be able to integrate information from multiple frames, updating temporal information in real-time. In the literature, many works provide use cases in which it is crucial to have an efficient and reliable MOT algorithm. These include:

\begin{figure}[t]
\centering
    \includegraphics[width=0.9\columnwidth]{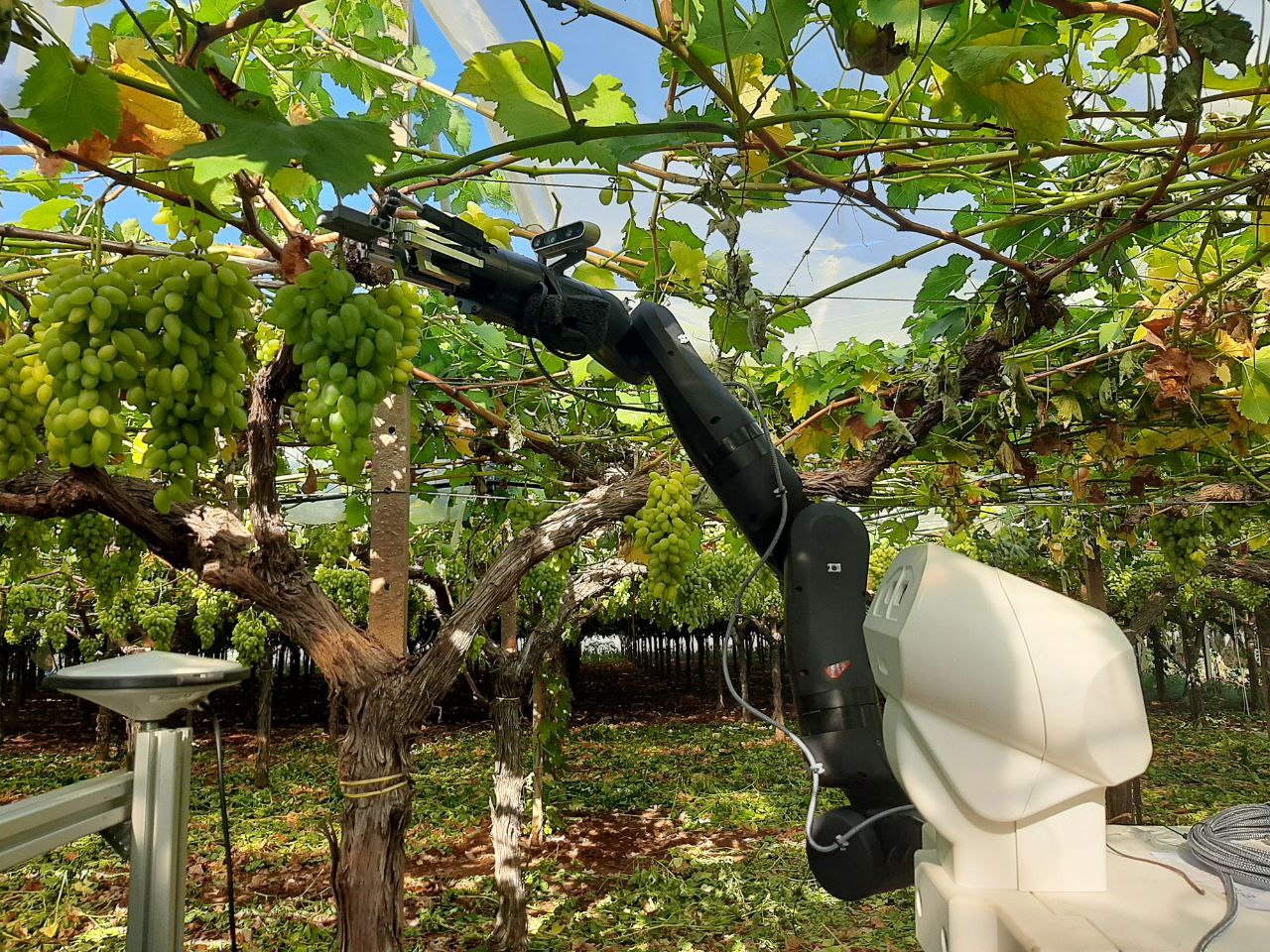}
    \caption{\footnotesize Agricultural robotic platform used in the EU Project CANOPIES for operations in table grape vineyards, equipped with an Intel RealSense d435i camera on the wrist. AgriSORT is implemented on this robot, and is used for tracking grapes.}
    \label{fig::robot}
\end{figure}

\begin{itemize}
    \item \textbf{Efficient use of resources:} By tracking individual plants, farmers can
    optimize resources such as water, fertilizer, and pesticides, saving costs
    and reducing the environmental impact of agriculture \cite{3_s21196657}.
    \item \textbf{Crop management}: MOT can be combined with other computer vision
    techniques in the field to make real-time decisions. Some of those include
    the identification of weeds to apply pesticides and herbicides \cite{4_applmech3030049}, precision spraying \cite{1_lettuceTrack}, and estimation of the quality of crops.
    \item \textbf{Yield estimation:} For yield estimation, tracking approaches are necessary to ensure objects are only counted once \cite{5_article}\cite{6_CIARFUGLIA2023107624}
    \item \textbf{Improved accuracy:} Multiple object tracking technology can help agricultural robots to accurately track each plant, even in challenging conditions, such as when plants are out of the camera's field of view \cite{2_farmMot}. In particular, tracking can provide more consistent results than detection-only applications for problems like quality assessment.
\end{itemize}

\begin{figure*}[t]
\footnotesize
\centering
\includegraphics[width=0.9\textwidth]{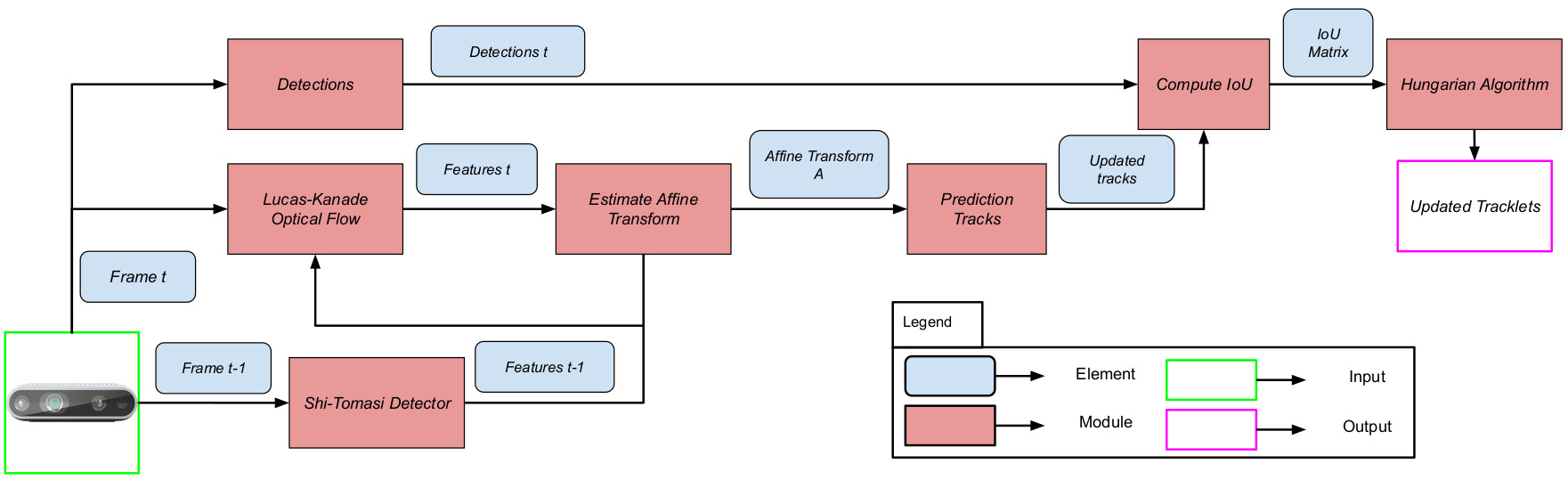}

\caption{\footnotesize Overview of our AgriSORT tracker pipeline. The process starts with the estimation of the relative camera motion. At first we extract features using the Shi-Tomasi method in the previous and current frame to compute the Optical Flow via the Lucas-Kanade algorithm. Using the generated matches we estimate an affine transform that expresses the motion of the camera. The estimated matrix is then used to propagate the state of the previous tracks in the current frame. At the end we compute the IoU with the current detection to perform the association and update the state of the tracklets.}
\label{fig::pipeline}
\end{figure*}

Multi-object tracking (MOT) is a computer vision problem that aims to identify and locate objects of interest in a video sequence, to associate them across frames to keep track of their movements over time. MOT is a fundamental issue for various applications, one of which is precision agriculture. Most SOTA tracking methods, such as SORT \cite{Bewley_2016}, DeepSORT \cite{wojke2017simple}, or JDE \cite{wang2020realtime}, use approaches based on appearance models since they are designed to track easily distinguishable moving objects from a still or slow-moving camera. While this approach is very effective in the context of classical MOT targets such as cars or people \cite{Geiger2012CVPR, MOTChallenge2015, MOTChallenge20}, for which it is possible to extract reliable appearance features, we found that they tend to be less effective when applied to the agricultural context. The main reason behind this is that in a farm environment, all crops are visually similar and static. In addition, appearance models are trained on additional data, besides those used to train the detection model, that is often unavailable in agricultural contexts, where data is scarce and often requires experts for labelling. To address these issues, we present AgriSORT, a perception pipeline incorporating the detection and tracking of crops, designed to be used on a robotic platform. Since our method does not use appearance features to identify instances in consecutive frames, we solve the association problem by using a Kalman Filter formulation different from the usual MOT algorithms, designed to follow the rapid camera movements with higher performance. In addition, the proposed approach is general and easily extendable to any crop type for which a working detector is available. The contribution of this work can be summarized as follows:

\begin{itemize}
	\item A MOT tracking solution especially suited for agricultural robotics scenarios is proposed. 
	\item A novel agricultural MOT benchmark, based on sequences collected on a table grape vineyard, is presented.
    \item The effectiveness of the proposed solution is experimentally evaluated and compared with the recent state-of-the-art MOT solutions.
\end{itemize}

\section{Related works}
\subsection{Vision-based application for precision agriculture}
Recently, robotic systems with different degrees of autonomy are spreading in agriculture to support farmers in various operations. LIDARs and GPS \cite{liu2023orb, underwood2015lidar} are among the most used sensors for these applications. However, these systems are costly and impose significant constraints from the point of view of the overall weight, the power used, and the availability of a fixed reference system on the ground. Therefore, the interest in vision strategies has grown over the years as they require only a camera of low weight and low cost. Many works in literature \cite{haug2014plant, lottes2017uav, milioto2017real} propose methods to solve the problem of detecting crops using hand-crafted features. However, they require careful tuning to achieve good domain generalization, so more recent approaches use the power of Deep Neural Network (DNN) detectors to detect crops. To be more specific, FasterRCNN has been deployed for the detection of apples \cite{mai2020faster}, mangoes \cite{bargoti2017deep}, tomatoes \cite{mu2020intact}, and oranges \cite{mai2020faster}. Other works employed the YOLOv3 \cite{liu2020yolo} architecture or the YOLOv5 for detecting tomatoes \cite{zhaoxin2022design}.
\subsection{Multiple Object Tracking}
In the MOT literature, methods are classified as separate and joint trackers. Separate trackers \cite{Bewley_2016, wojke2017simple, du2023strongsort, zhang2022bytetrack, du2021giaotracker, cao2023observationcentric, zhang2021fairmot, 1_lettuceTrack} follow the tracking-by-detection paradigm, which localizes and classifies all targets first and then associates detections belonging to the same object using information on appearance, motion, or both. On the other hand, joint trackers \cite{bergmann2019tracking, wang2020realtime, zhou2020tracking, tokmakov2021learning, sun2020transtrack, meinhardt2022trackformer, li2022simpletrack, lu2020retinatrack} unify the detection and appearance model into a single framework and train them together. Compared to separate trackers, the main advantage of joint methods is their high inference speed and comparable performance in simple scenes. However, they tend to fail in more sophisticated scenarios. The closest work in motivation to ours is LettuceTrack \cite{1_lettuceTrack}, which proposes a tracking pipeline for the detection of lettuce using a wheeled robotic platform with an RGB camera pointed downwards, while the robot is moving in a straight line. The association problem is solved with geometrical considerations by exploiting the regular pattern of the planted crops. 
For these reasons, their approach is not effective for unconstrained motion and variable illumination conditions, typical of robotic monitoring applications, while our approach has a much wider applicability. We demonstrate it on the case of table grape cultivation, which has specific challenges of instance similarity, instances distance and separability, and rapid illumination changes.

\section{Materials and Methods}
\label{sec::methods}
In this section we describe the data collection and give a detailed description of the tracking pipeline. In the following section we give an overview of the complete pipeline, then in Section \ref{sec:data} the experimental field and the data collection and labelling are described. In Sections \ref{sec:kalman} and \ref{sec::association} we discuss the Kalman Filter formulation and the association process. The algorithm pipeline is presented in Fig \ref{fig::pipeline}.

\begin{figure}[t]
\centering
    \vspace{5pt}
    \begin{subfigure}[]{0.13\textwidth}
        \includegraphics[width=\columnwidth]{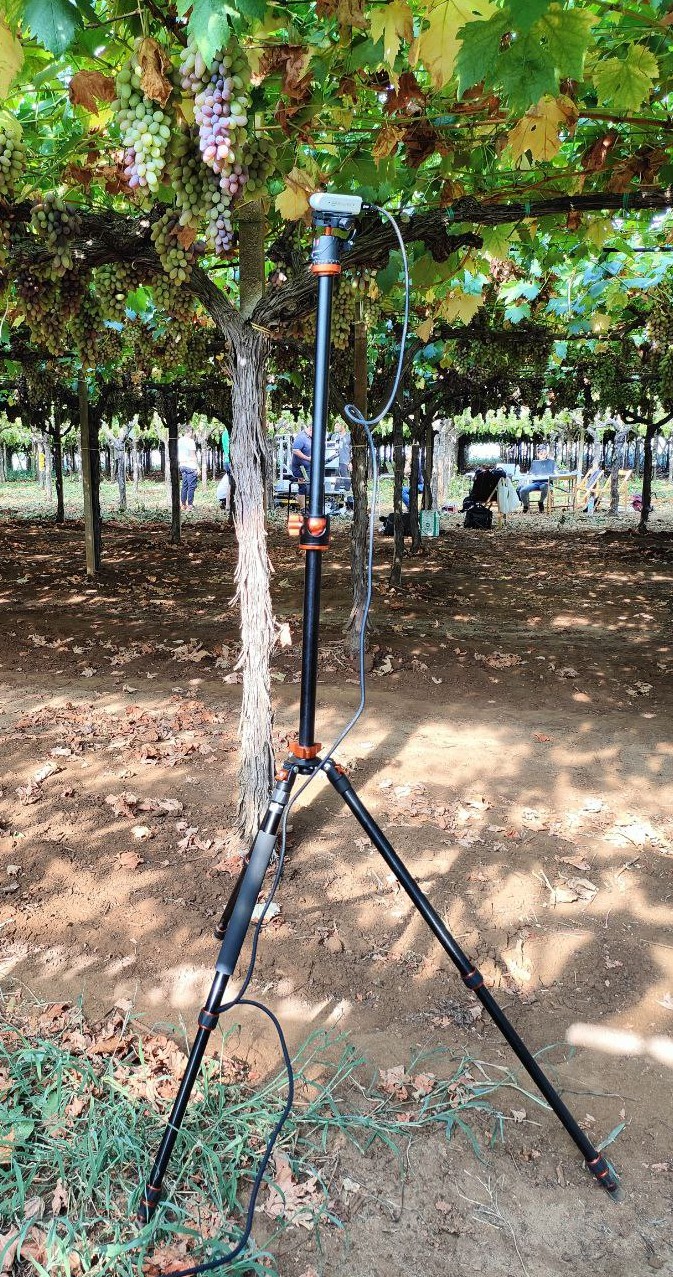}
    \end{subfigure}
        \begin{subfigure}[]{0.329\textwidth}
        \includegraphics[width=\columnwidth]{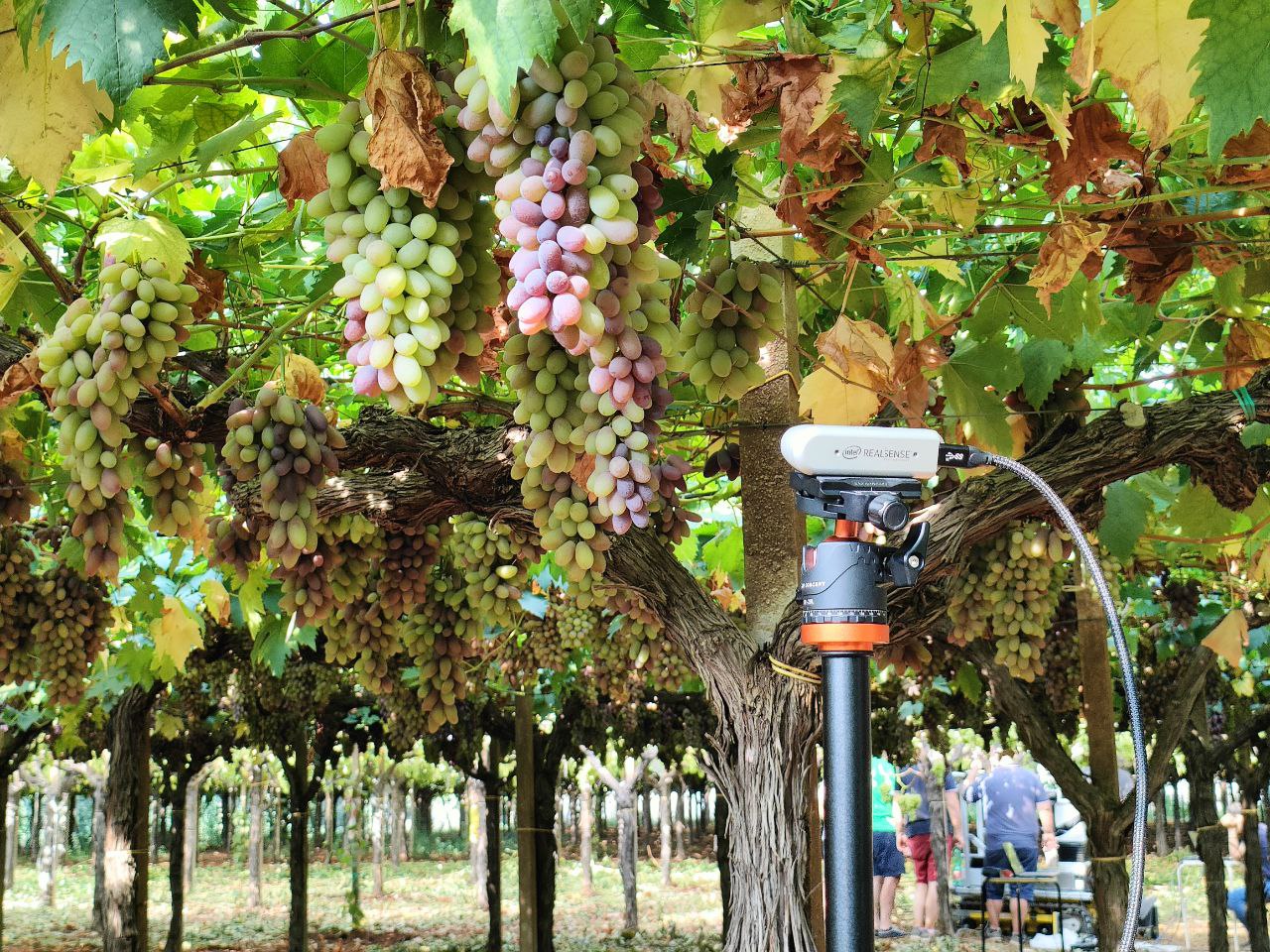}
    \end{subfigure}
    \caption{\footnotesize Experimental setup for data acquisition, a Intel RealSense d435i camera mounted on a tripod for stabilization during movements.}
    \label{fig::setup}
\end{figure}

\subsection{Data acquisition} \label{sec:data}
The dataset used in this paper has been collected in a vineyard of table grapes located in southern Lazio (Italy). The vineyards are structured as a traditional trellis system called Tendone, with a vast distance between each plant of 3 meters. Plantations are all older than three years, so they are in full production and health, thus representing a typical working condition for validating agronomic activities. We used a depth camera Intel RealSense D435i (Fig. \ref{fig::setup}) to collect the data while moving along the vineyard rows, pointing towards the grape bunches. Even though the full resolution of the camera is FullHD (1920x1080), we recorded the data with HD resolution (1280x720) at 30 FPS, to allow for alignment between depth and RGB images. We downsampled some sequences to 10 FPS to see how the detector works at a lower framerate. During the data acquisition campaign, the images are collected with the handheld camera, simulating the motion of the robot during operations like harvesting or yield estimation. We manually annotated the collected data to provide a test set to validate the performance of the proposed method using CVAT (Computer Vision Annotation Tool) \cite{CVAT_ai_Corporation_Computer_Vision_Annotation_2022} with the MOT format.

\vspace{10pt}
\begin{table}[htbp]
\centering
\caption{\footnotesize List of sequences acquired and labelled for MOT with their characteristics}
\begin{adjustbox}{width=0.45\textwidth}
\begin{tabular}{|c|c|c|c|c|}
\hline
Dataset & CloseUp1 & CloseUp2  & Overview1 & Overview2\\
\hline
Resolution & 1280x720 & 1280x720 & 1280x720 & 1280x720\\ 
Length (frames) & 300 & 300 & 100 & 100\\
FPS & 30 & 30 & 10 & 10\\ 
Tracks & 23 & 22 & 20 & 31\\ 
Boxes & 2583 & 3581 & 1040 & 721\\ 
\hline
\end{tabular}
\end{adjustbox}
\label{tab:dataset}
\end{table}

\begin{figure*}[t]
\centering
        \vspace{5pt}
	\begin{subfigure}[]{0.45\textwidth}
		\includegraphics[width=\columnwidth]{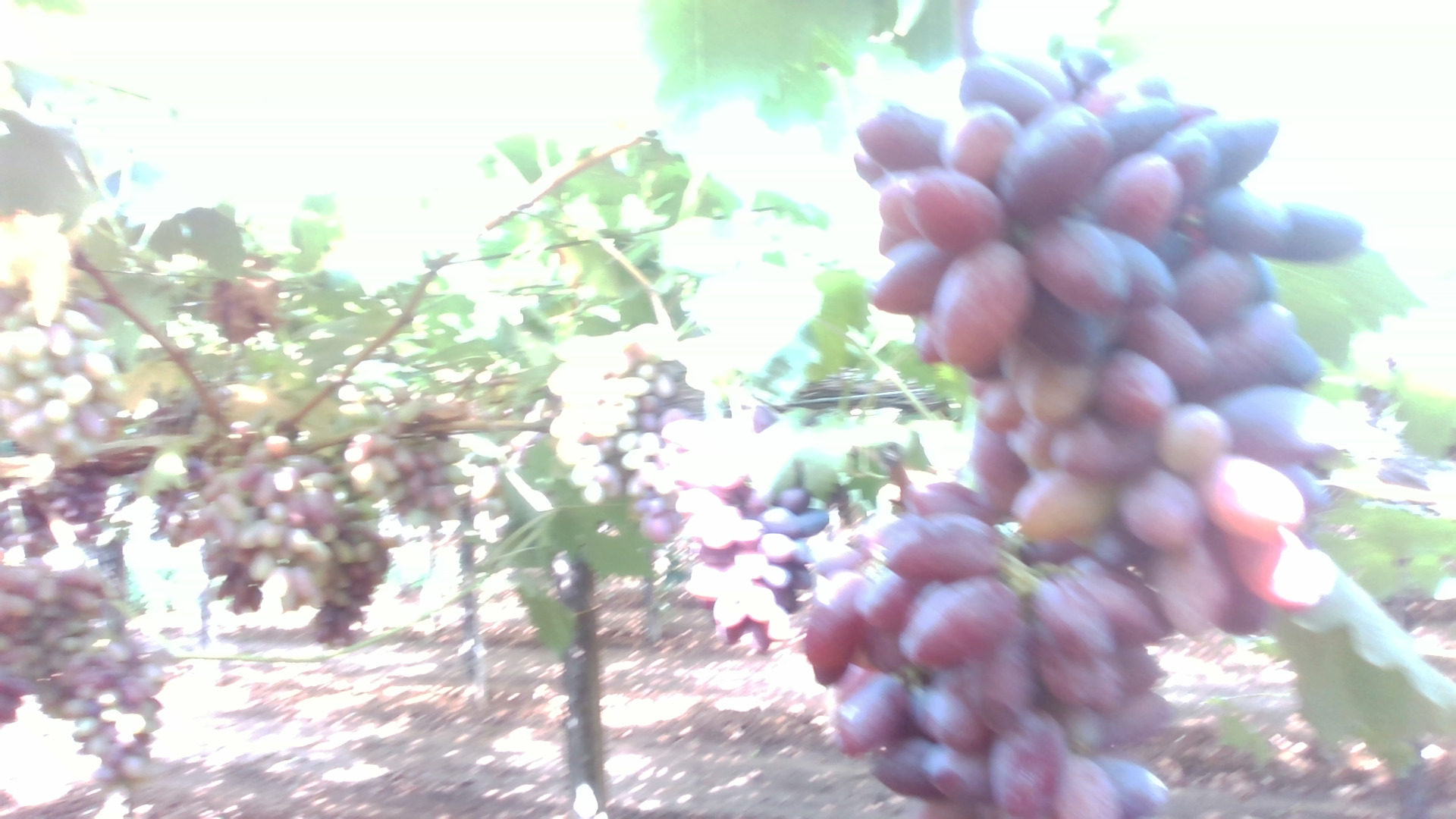}
		\caption{}
	\end{subfigure}
        \begin{subfigure}[]{0.45\textwidth}
		\includegraphics[width=\columnwidth]{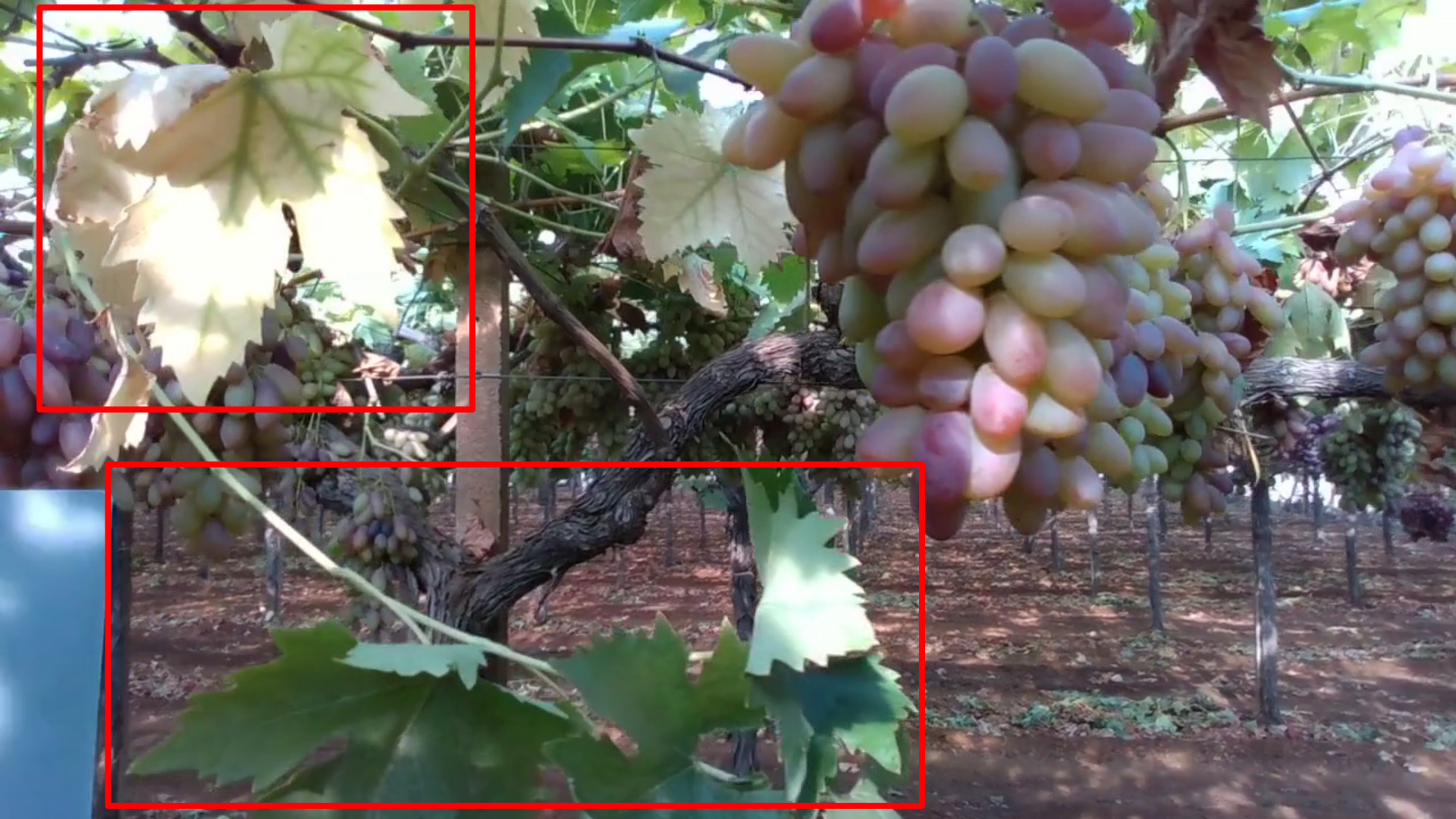}
		\caption{}
	\end{subfigure}
	\caption{\footnotesize Example of difficult cases present in the experimental field. Strong frontal illumination and motion blur due to fast motion (a), and occlusions due to leafs and branches (b).}
\label{fig:difficulties}
\end{figure*}

The sequences involve various movements within the vineyard rows, such as sudden changes in direction, U-shaped patterns, and close-up shots of table grape bunches. Details of the dataset are summarized in Table \ref{tab:dataset}. The analysis of table grapes for the purposes of robotic agriculture presents intriguing but formidable challenges due to the relatively unstructured nature of the environment. Unlike crops such as lettuce or cabbage that are planted in fixed patterns, table grapes grow along rows without any specific arrangement. This poses a challenge for tracking, as occlusions occur frequently due to overlapping crops, dense foliage, and poles. The dataset for this tracking task is unique, as the objects being tracked remain static while the camera moves relatively to the crops. This task is made difficult by irregular and unpredictable (human or robotic) motion, which can be very fast. Additionally, the agricultural environment is highly unstructured, and the terrain is steep, resulting in noise from camera vibrations and minor oscillations. Extreme movements due to potholes or other obstructions also cause motion blur. Illumination changes rapidly and can sometimes be directed toward the camera, causing unusable frames. Some examples of these difficulties can be found in Fig. \ref{fig:difficulties}. 

\subsection{Kalman Filter formulation} \label{sec:kalman}

Each of the detected objects is instantiated with a Kalman Filter (KF) that tracks its geometrical properties through time.
There are a number of facts to consider when designing the KF for bounding box tracking in the context of agricultural scenarios. Since robot motion along the field happens on a rough terrain, the camera is subject to huge vibrations that translate to unpredictable motion in the image coordinates. It is common practice in MOT to include in the KF state vector the derivatives of the bounding box size and motion, and this works well when the hypothesis of linear motion holds. Since this is not the case, the derivative of motion and size become detrimental to the KF estimation. 
A second consideration is that, given camera high frame rates compared to its velocity, it is easy to detect rapid direction changes and to approximate the camera motion using simple affine transformations. Therefore, relying on the motion vector for correction of the state vector, instead of using a transition matrix of any sort, is a simple and effective solution to the tracking problem in this context. 

The state and measurement vectors are defined as follows:

 \begin{equation}
\label{eq::state}
x_k = [x_{c}(k), y_{c}(k), w(k), h(k)]^T
\end{equation}

\begin{equation}
\label{eq::measurement}
z_k = [z_{x_c}(k), z_{y_c}(k), z_{w}(k), z_{h}(k)]^T
\end{equation}
where $k$ is the time step, $(w,h)$ are the width and height of the bounding box, $\mathbf{x}$ is the state vector, and $\mathbf{z}$ is the measurement vector. 

To estimate the camera motion we use Lucas-Kanade sparse optical flow correspondences \cite{lucas1981iterative} and solve for an affine transformation matrix, defined as follows:

\begin{equation}
\label{eq::affine}
A^{k}_{k-1} = 
    \begin{bmatrix}
    a11 & a12 & a13\\
    a21 & a22 & a23
    \end{bmatrix}
\end{equation}
The affine transform parameters represent a linear mapping between points in the reference and current images, describing how to transform the reference image to align it with the current image. In particular the translation components $(a_{13}, a_{23})$ provide an estimate of the camera's displacement in the $x$ and $y$ directions, while the scale and rotation components $(a_{11}, a_{12}, a_{21}, a_{22})$ can indicate changes in scale and rotation.

 Affine motion can capture translation, rotation, scaling, and shearing effects but not perspective distortions. Since most of the robot motion is parallel to the fruit orchard rows, the perspective effects are limited. In addition, the high frame rate further reduces the error due to this approximation. The performance advantages of using simple affine transformation will be discussed in Section \ref{sec::experiments}. Given the affine transform, the custom prediction step updates only the center of the bounding box, projecting it between frames and keeping width and height fixed. Therefore the propagation step follows Eq.(\ref{eq::propagation}).

\begin{equation}
\label{eq::propagation}
\begin{bmatrix}
\tilde{x_{c}(k)}\\
\tilde{y_{c}(k)}\\
\tilde{w(k)}\\
\tilde{h(k)}
\end{bmatrix}
=
\begin{bmatrix}
a11 \cdot x_{c}(k-1) + a12 \cdot y_{c}(k-1) + a13\\
a21 \cdot x_{c}(k-1) + a22 \cdot y_{c}(k-1) + a23\\
w(k-1)\\
h(k-1)
\end{bmatrix}
\end{equation}

\subsection{Association}
\label{sec::association}
For what concerns the association step, given a set of observations at frame $t$ and a set of predicted observations according to the estimation model, we aim to find the optimal assignment of observations to predicted states. 

We formalize the problem as an optimization problem over an assignment matrix $A$, where each element ($A_{ij}$) indicates the intersection over union (IOU) between the $i_{th}$ observation and the $j_{th}$ predicted state. Specifically, $A_{ij} = 1$ if the $i_{th}$ observation perfectly overlaps with the $j_{th}$ prediction, while $A_{ij} = 0$ means the two objects do not overlap. To find the optimal assignment between observations and predicted states that minimizes the overall IOU, we use the Hungarian algorithm \cite{kuhn1955hungarian}, a well-known solution for these problems. 

\section{Experiments}
\label{sec::experiments}

\begin{figure*}[t]
\centering
    \begin{subfigure}[]{0.24\textwidth}
        \centering
        \caption*{\textbf{AgriSORT}}
        \includegraphics[width=\columnwidth]{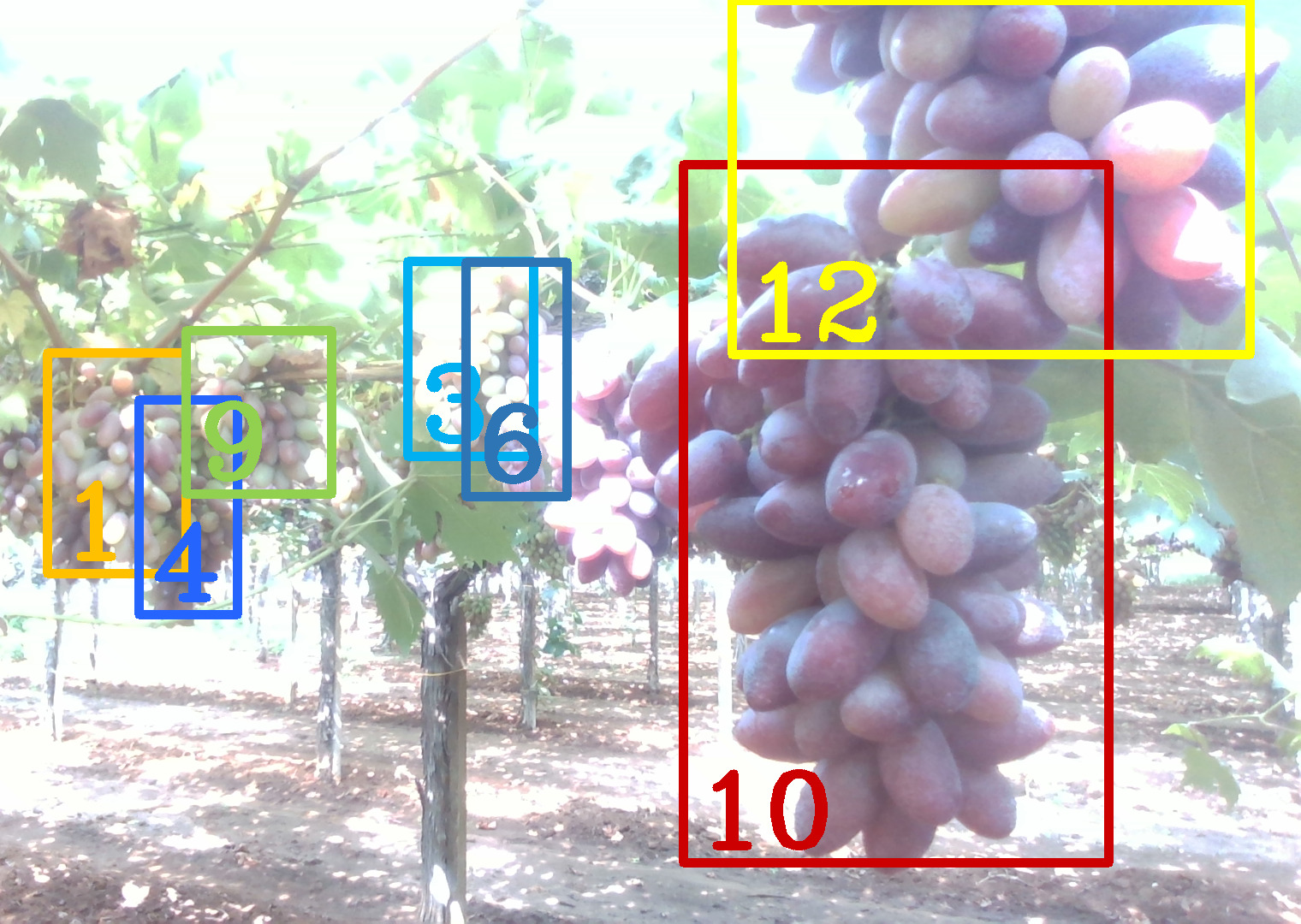}
    \end{subfigure}
    \begin{subfigure}[]{0.24\textwidth}
        \centering
        \caption*{\textbf{SORT}}
        \includegraphics[width=\columnwidth]{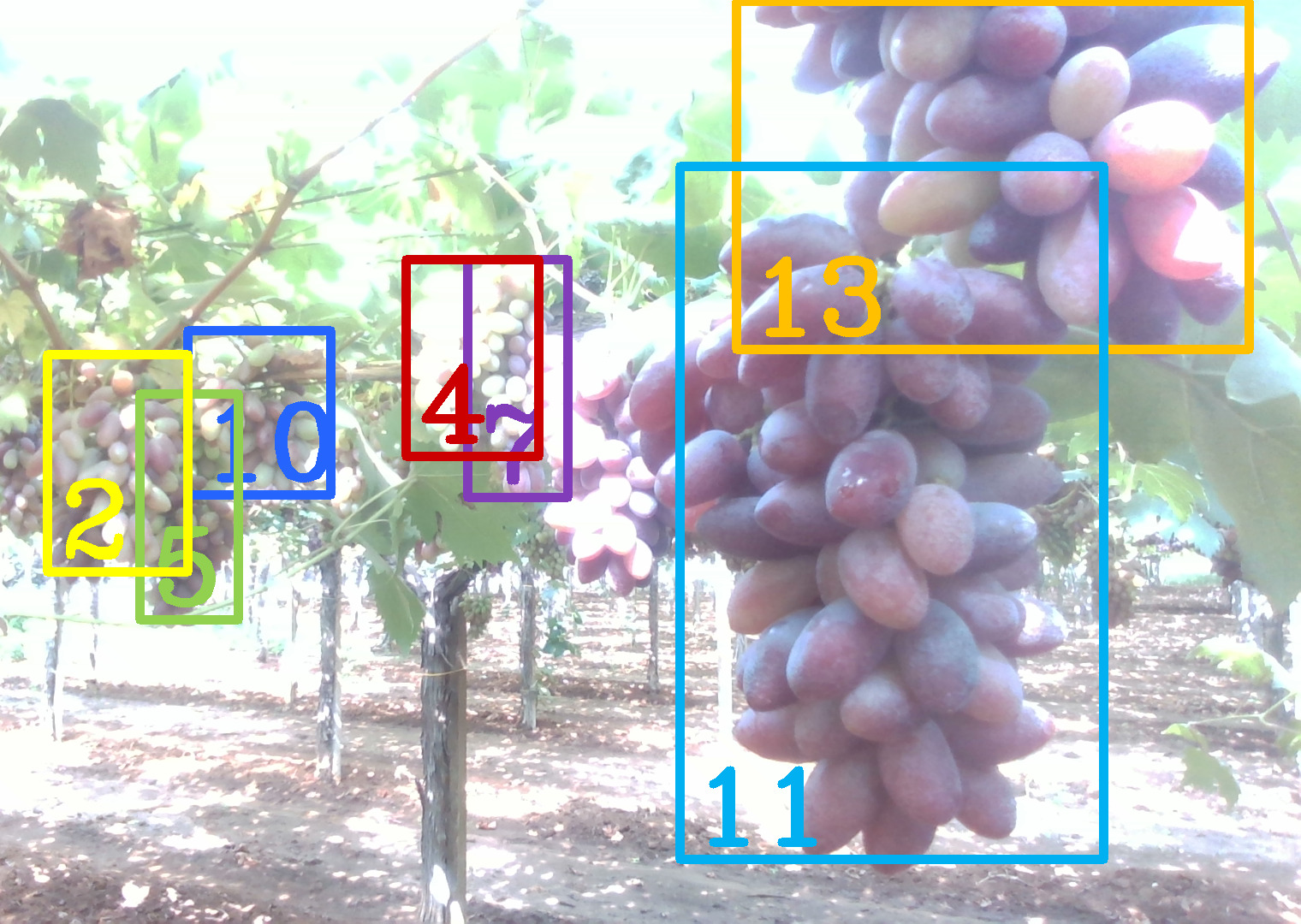}
    \end{subfigure}
    \begin{subfigure}[]{0.24\textwidth}
        \centering
        \caption*{\textbf{OC-SORT}}
        \includegraphics[width=\columnwidth]{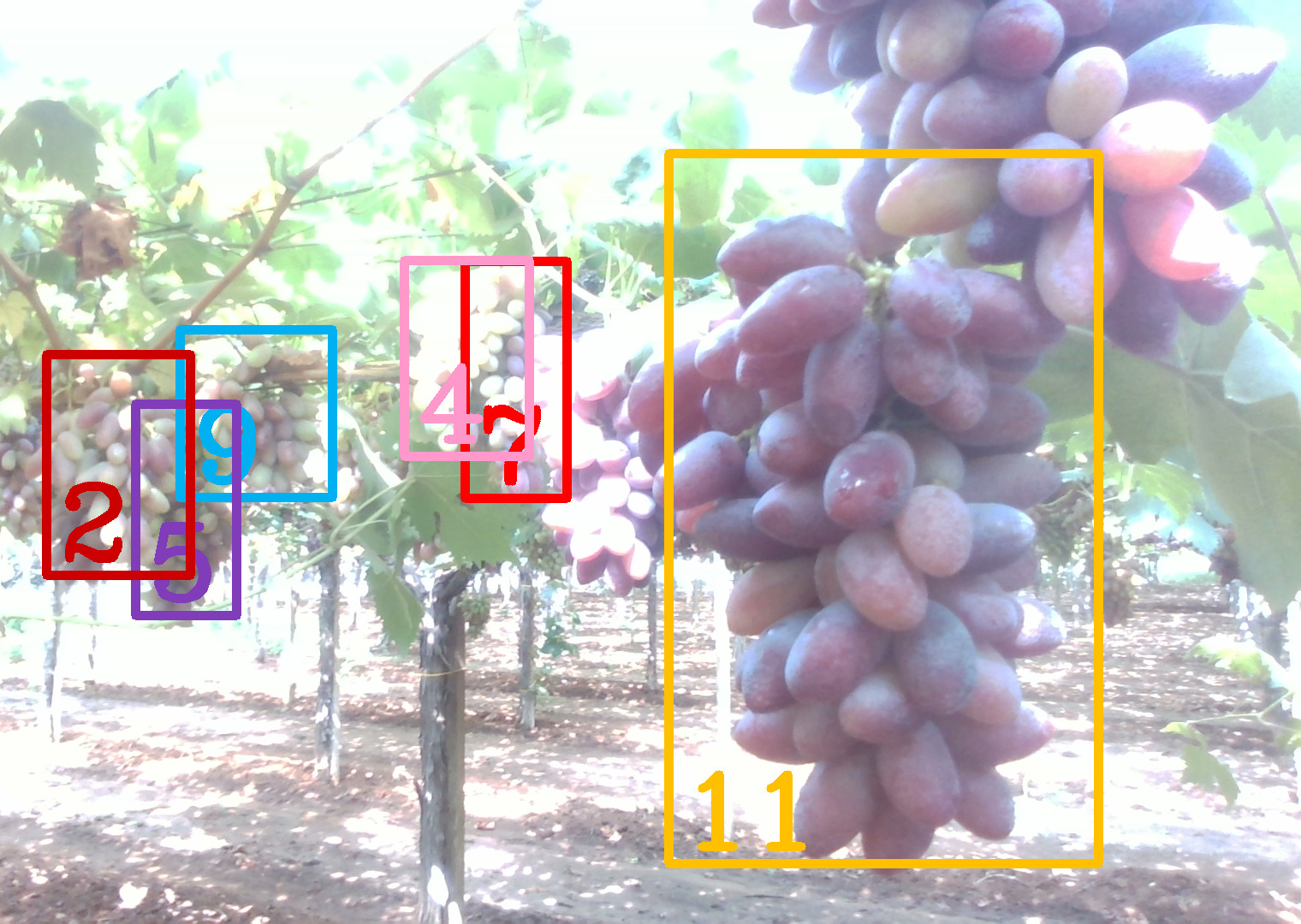}
    \end{subfigure}
    \begin{subfigure}[]{0.24\textwidth}
        \centering
        \caption*{\textbf{BYTE}}
        \includegraphics[width=\columnwidth]{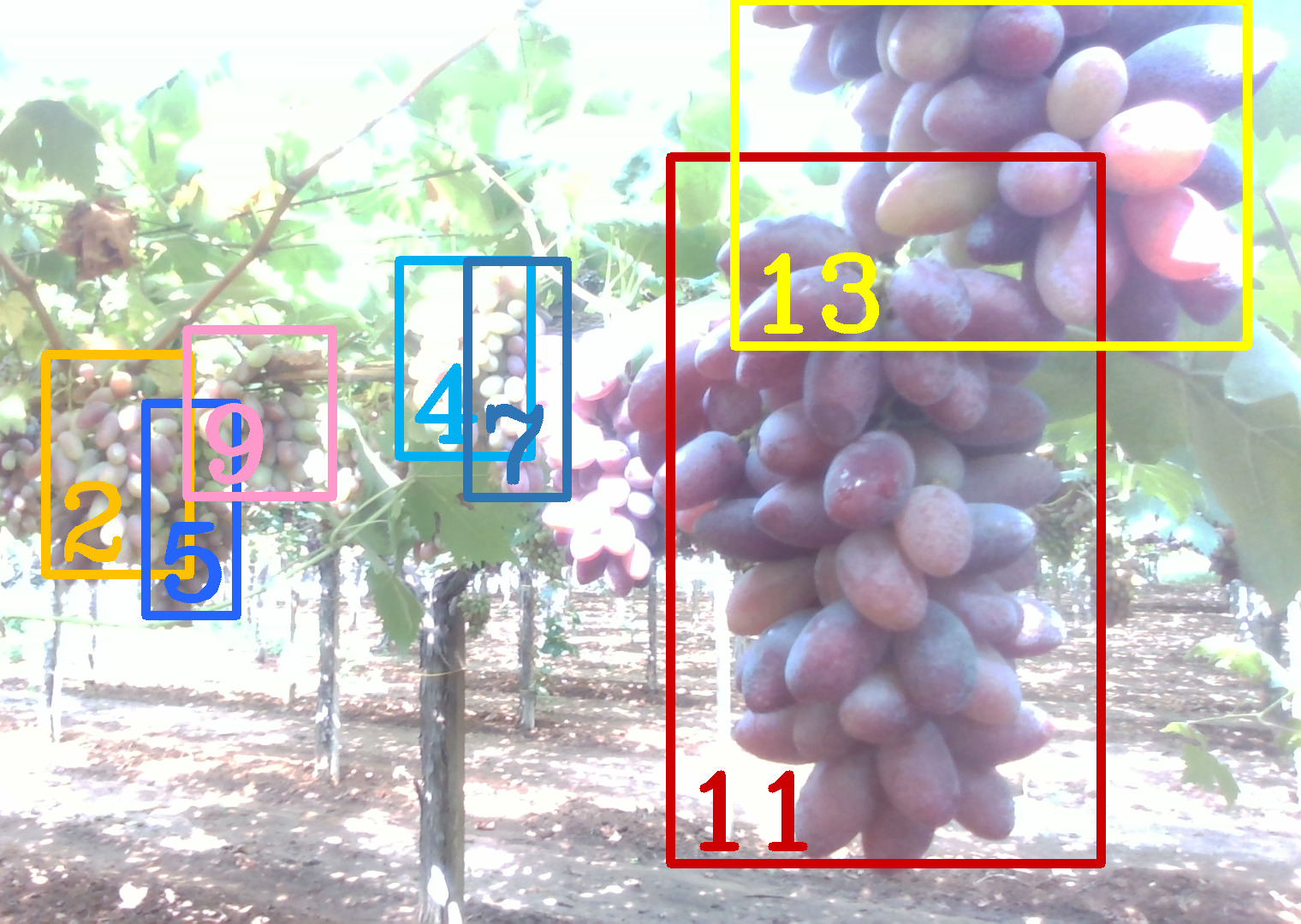}
    \end{subfigure}

    \vspace{5pt}
    \begin{subfigure}[]{0.24\textwidth}
        \includegraphics[width=\columnwidth]{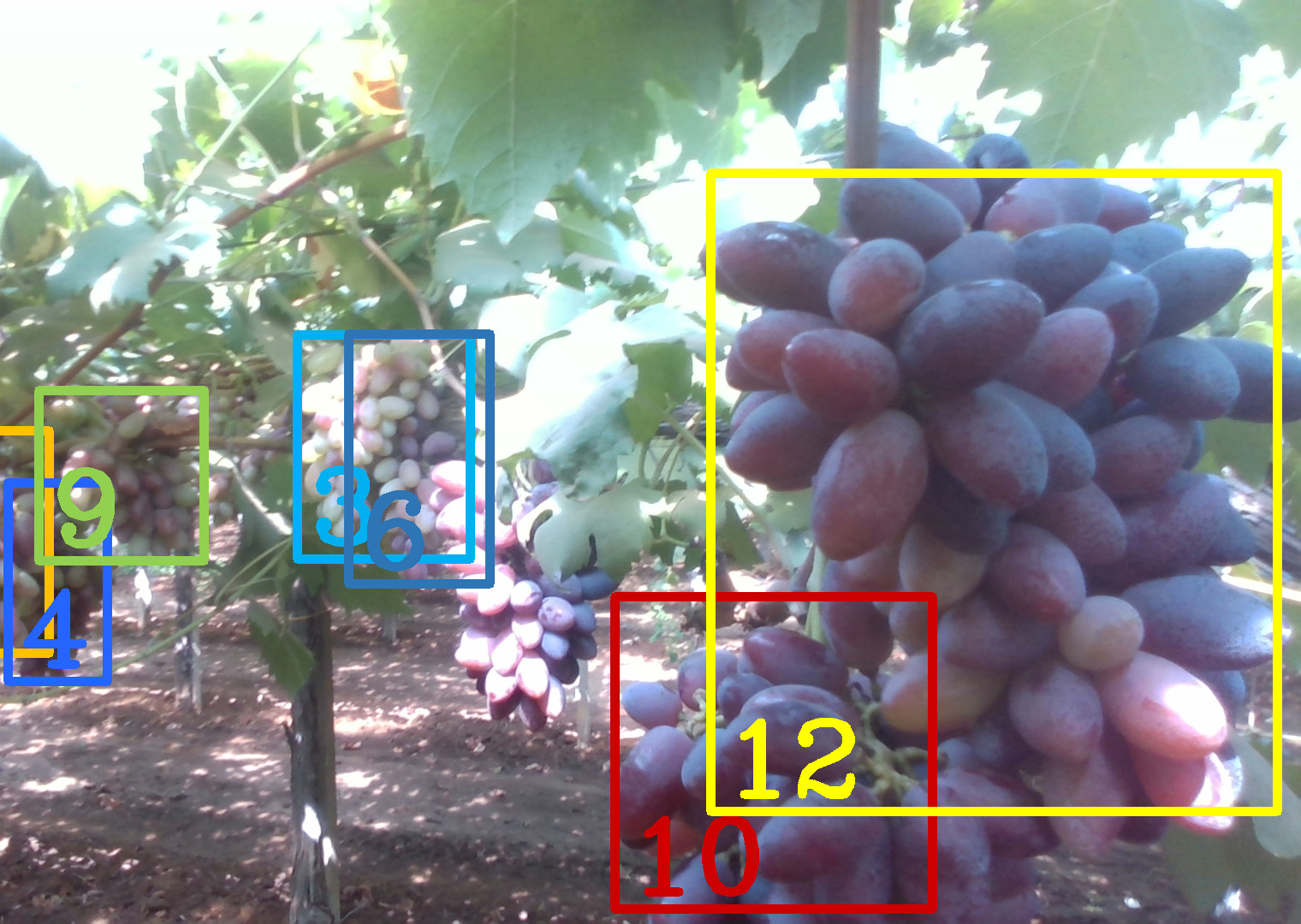}
        \caption{}
    \end{subfigure}
    \begin{subfigure}[]{0.24\textwidth}
        \includegraphics[width=\columnwidth]{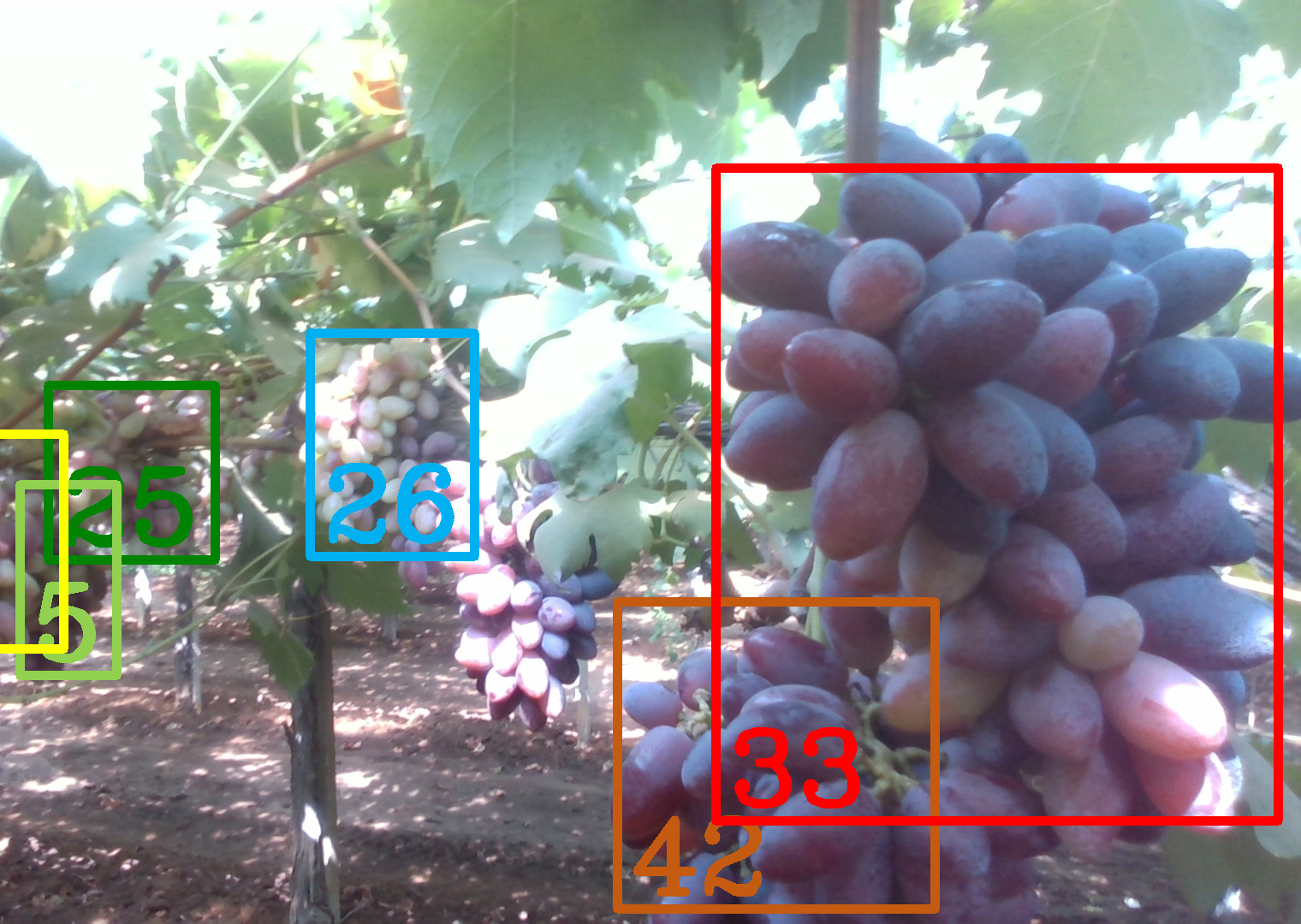}
        \caption{}
    \end{subfigure}
    \begin{subfigure}[]{0.24\textwidth}
        \includegraphics[width=\columnwidth]{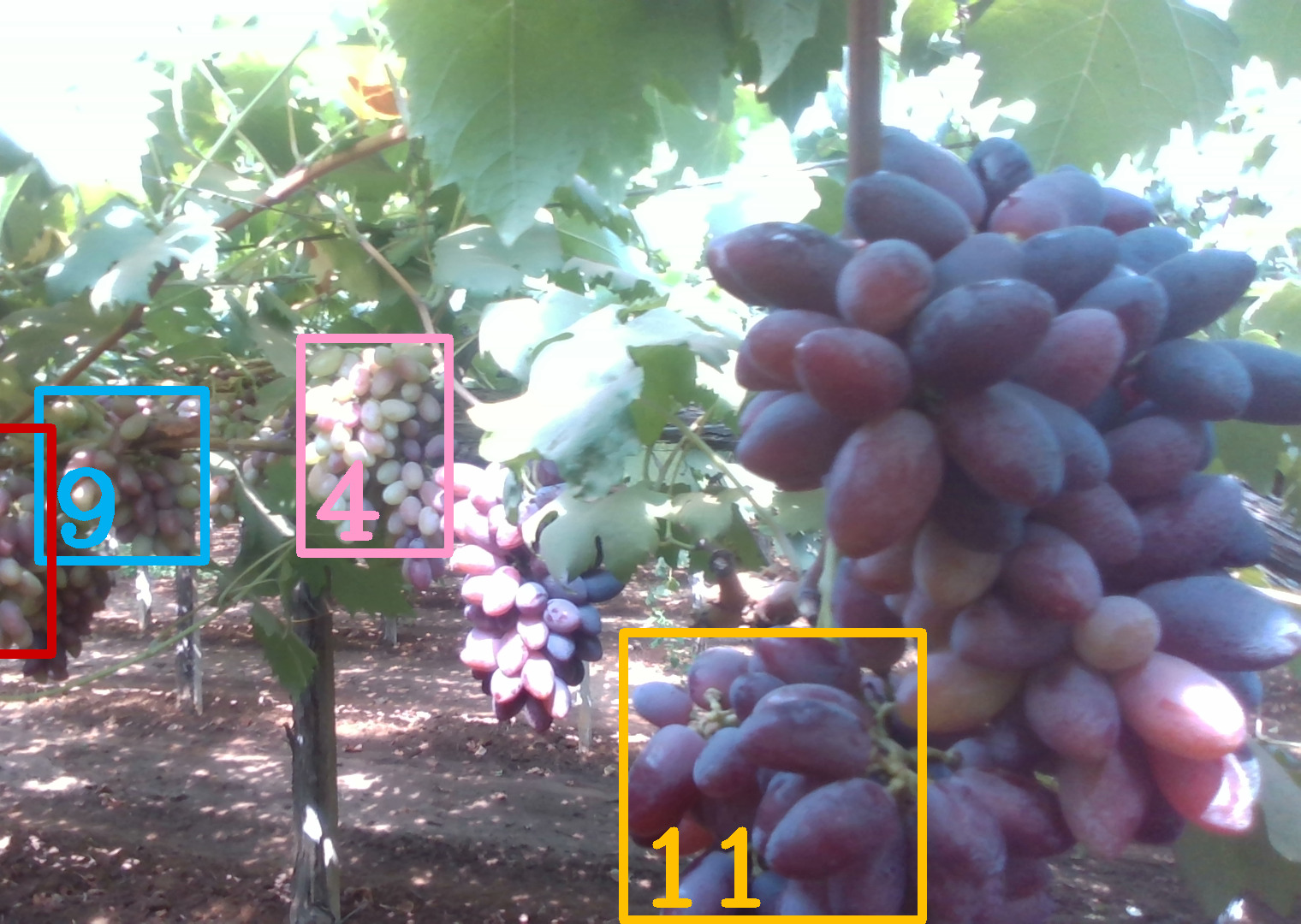}
        \caption{}
    \end{subfigure}
    \begin{subfigure}[]{0.24\textwidth}
        \includegraphics[width=\columnwidth]{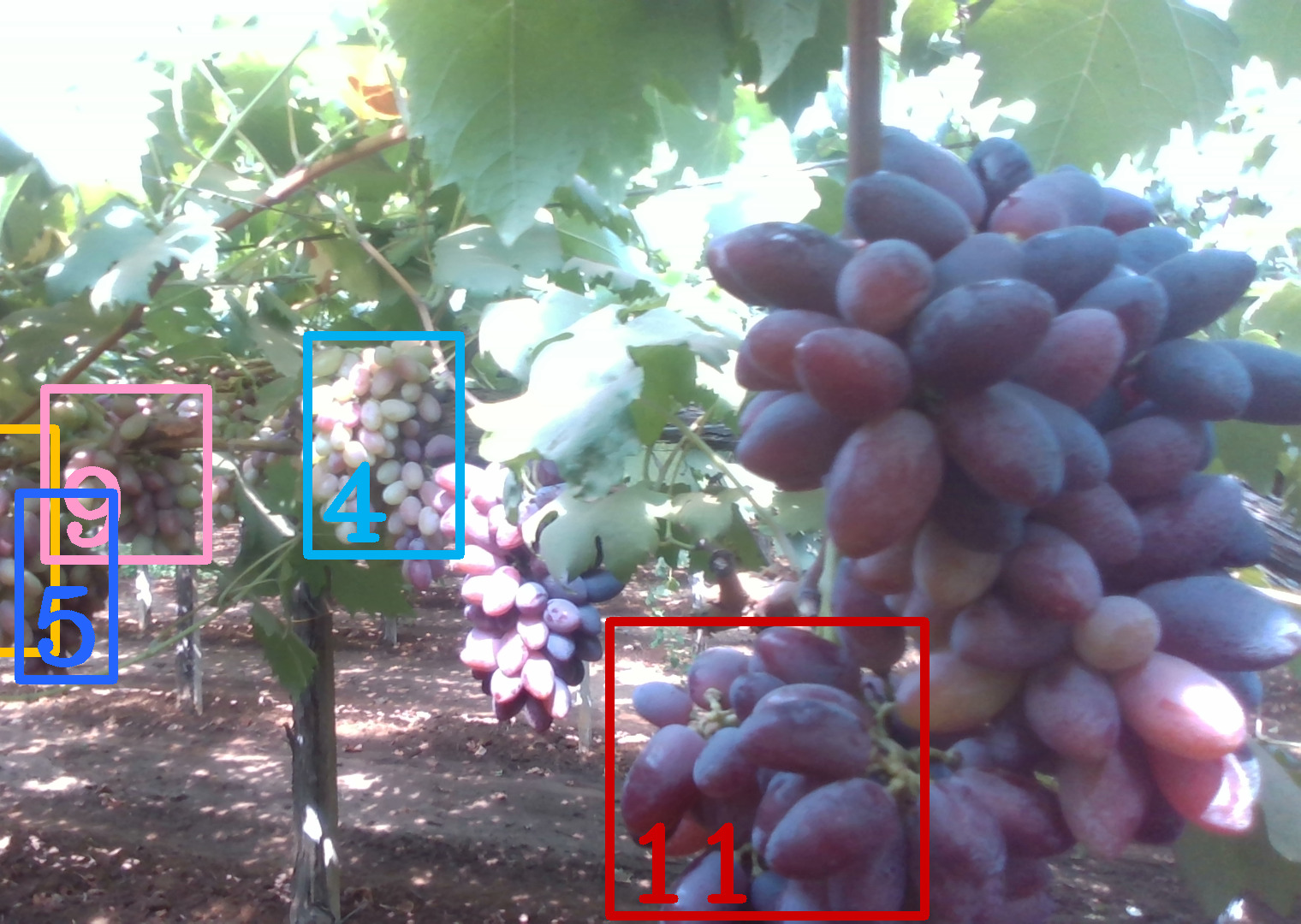}
        \caption{}
    \end{subfigure}
    \caption{\footnotesize Qualitative evaluation of the performance of Agrisort and other SOTA trackers in the "CloseUp1" sequence at frame 20 (top row) and 80 (bottom row). AgriSORT (a) is able to keep consistency on most tracks, without switching IDs and providing the least number of false negatives. SORT (b) suffers in this sequence, it loses almost all the IDs, due to the presence of fast non-linear motion and illumination changes. OC-SORT (c) and BYTE (d) perform better than SORT, in particular without switching IDS, however their performance is far from AgriSORT because they lose some of the tracks both in foreground and background.}
    \label{fig::comparison}
\end{figure*}

In the following section, we explain the implementation details of the proposed method as well as the evaluation metrics to test the performance and the comparison with other state-of-the-art MOT methods.

\begin{table*}[ht]
\centering
\footnotesize
\caption{\footnotesize Comparison of AgriSORT performance with other tracking-by-detection trackers on agricultural sequences}
\begin{adjustbox}{width=0.8\textwidth}
\begin{tabular}{ |c|c|c|c|c|c|c|c|c|c|c|} 
    \hline
    Sequence & Method & MOTA↑ & IDF1↑  & HOTA↑ & FP↓ & FN↓ & IDs↓ & MT↑ & ML↓\\
    \hline
    CloseUp1 & BOTSORT \cite{aharon2022bot} & 48.66 & 72.72 & 40.51 & 197 & 1381 & 56 & 4 & 7\\ 
    & BYTETRACK \cite{zhang2022bytetrack} & 52.19 & 56.01 & 37.153 & 51 & 1157 & 27 & 4 & 7\\
    & OCSORT \cite{cao2023observationcentric} & 48.70 & 62.27 & 40.52 & 37 & 1272 & 16 & 2 & 7\\ 
    & SORT \cite{Bewley_2016} & 48.238 & 44.27 & 31.43 & 31 & 1271 & 35 & 2 & 7\\ 
    & StrongSORT \cite{du2023strongsort} & 57.37 & 60.25 & 41.20 & \textbf{21} & 1054 & 26 & 6 & 7\\ 
    & AgriSORT (ours) & \textbf{65.93} & \textbf{72.00} & \textbf{48.71} & 459 & \textbf{872} & \textbf{11} & \textbf{11} & \textbf{4}\\
    \hline

    CloseUp2 & BOTSORT \cite{aharon2022bot} & 51.58 & 61.645 & 47.02 & 405 & 1805 & 30 & 8 & 4 \\
        & BYTETRACK \cite{zhang2022bytetrack} & 53.81 & 64.42 & 46.16 & 371 & 1703 & 18 & 8 & 3 \\
        & OCSORT \cite{cao2023observationcentric} & 48.42 & 62.95 & 45.99 & 233 & 1829 & \textbf{9} & 6 & 5 \\
        & SORT \cite{Bewley_2016} & 47.11 & 55.41 & 43.48 & 424 & 2046 & 38 & 6 & 6 \\
        & StrongSORT \cite{du2023strongsort} & 56.46 & 55.19 & 43.16 & \textbf{371} & 1703 & 18 & 8 & 3 \\
        & AgriSORT (ours) & \textbf{66.13} & \textbf{73.00} & \textbf{56.08} & 655 & \textbf{1146} & 10 & \textbf{12} & \textbf{1} \\
    \hline
    Overview1 & BOTSORT \cite{aharon2022bot} & 48.46 & 64.77 & 40.32 & 244 & 425 & 53 & 5 & 1 \\
        & BYTETRACK \cite{zhang2022bytetrack} & 55.58 & 62.94 & 41.48 & 248 & 448 & 29 & 7 & 2 \\
        & OCSORT \cite{cao2023observationcentric} & 53.94 & 69.43 & 42.24 & 143 & 411 & 15 & 5 & 2 \\
        & SORT \cite{Bewley_2016} & \textbf{62.69} & 73.91 & 51.20 & \textbf{108} & 367 & \textbf{9} & 8 & 2 \\
        & StrongSORT \cite{du2023strongsort} & 48.94 & 62.29 & 38.81 & 346 & 413 & 22 & 8 & 0 \\
        & AgriSORT (ours) & 62.21 & \textbf{73.72} & \textbf{52.74} & 255 & \textbf{284} & 12 & \textbf{13} & \textbf{0} \\
    \hline
    Overview2 & BOTSORT \cite{aharon2022bot} & 30.652 & 50.193 & 32.434 & 249 & 396 & 55 & 8 & 6 \\
        & BYTETRACK \cite{zhang2022bytetrack} & 42.302 & 54.784 & 35.423 & 220 & 366 & 20 & 10 & 5 \\
        & OCSORT \cite{cao2023observationcentric} & 41.748 & 55.783 & 34.578 & 158 & 381 & \textbf{11} & 6 & 5 \\
        & SORT \cite{Bewley_2016} & \textbf{49.515} & \textbf{61.96} & \textbf{44.431} & \textbf{110} & 348 & 12 & 10 & 8 \\
        & StrongSORT \cite{du2023strongsort} & 34.258 & 54.093 & 34.28 & 304 & 341 & 17 & 12 & 6 \\
        & AgriSORT (ours) & 45.08 & 56.88 & 41.33 & 298 & \textbf{316} & 14 & \textbf{16} & \textbf{4} \\
    \hline
    
\end{tabular}
\end{adjustbox}
\label{tab:performance}
\end{table*}

\subsection{Implementation details}
The detector employed for this application is the YOLO-V5s model of the YOLO-V5 family because it provides the best trade-off between inference speed and accuracy, which is vital for real-time tracking applications. We trained the model on the table-grapes dataset presented in \cite{ciarfuglia2023weakly, ciarfuglia2022pseudo}, which consists of 242 annotated images and 1469 images automatically annotated using a pseudo-label generation strategy. The training details can be found in \cite{ciarfuglia2023weakly, ciarfuglia2022pseudo}. The NVIDIA GeForce RTX 3070 Ti Laptop GPU is the inference and training hardware. The choice for the system and measurement noise matrices $Q_k$ and $R_k$, respectively, consist of 4x4 diagonal matrices with fixed noise factors, which are $\sigma_q = 0.05$ and $\sigma_r = 0.00625$, also multiplied by a factor ($\delta t$) that depends on the framerate of the camera sequence ($0.033$ for the $30$ FPS sequences, and $0.1$ for the $10$ FPS sequences)

\begin{equation}
\begin{split}
\label{eq::matrices}
Q_k = diag_{4x4}((\sigma_q)^{2}) \cdot \delta t\\
R_k = diag_{4x4}((\sigma_r)^{2}) \cdot \delta t\\
\end{split}
\end{equation}

\subsection{Baselines}
To compare with other methods, we choose some baseline SOTA MOT approaches. In particular, we choose SORT because it is the first and most straightforward tracking-by-detection approach. Additionaly, we selected ByteTrack and OC-SORT because of their methodological solid innovations and the integration of a camera motion compensation module provided by OpenCV into the pipeline. We also compare with StrongSORT and BoTSORT since both achieved excellent results on pedestrian datasets MOT17 and MOT20. We shared the same detector model for each tracker to provide a fair comparison with the other state-of-the-art trackers. However, some methods (StrongSORT and BoTSORT) require an additional model for appearance feature extraction besides the detector, which we cannot finetune due to the absence of data. Therefore the pre-trained default models provided by the authors are used.

\subsection{Evaluation metrics}
Evaluations were performed according to a combination of widely accepted metrics defined by \cite{milan2016mot16, luiten2021hota, bernardin2008evaluating}. Those include Multiple Object Tracking Accuracy (MOTA), False Positive (FP), False Negative (FN), ID Switch (IDs), IDF1, Higher-Order Tracking Accuracy (HOTA), Mostly Tracked (MT), and Mostly Loss (ML). MOTA is a metric commonly used to assess the performance of MOT algorithms. It measures the overall tracking accuracy by considering false positives (FP) and false negatives (FN). MOTA focuses more on detection performance because the amount of FP and FN are more significant than the IDs. On the other hand, IDF1 is a metric that evaluates the ability of the tracker to correctly identify individual objects across frames, combining precision and recall of identity assignments. HOTA is an extension of MOTA that incorporates additional metrics to evaluate the tracking performance at different levels, including localization, classification, and association. FP refers to the number of objects incorrectly detected as present when they are not, while FN refers to the number of targets missed or not detected by the tracking algorithm when they are present in the scene. MT and ML are binary metrics indicating whether an object is mostly tracked or lost throughout the sequence. An object is considered mostly tracked (MT) if its overlap with the ground truth is above a certain threshold for a significant portion of its lifespan. To compute the above metrics, we use TrackEval \cite{luiten2020trackeval}, the official reference implementation for the HOTA metrics.

\subsection{Results and discussions}
The performance of the proposed method and compared to the baselines are summarized in Table \ref{tab:performance}. It can be seen that on the \textit{CloseUp} sequences our method always outperforms all the others. The fact that those sequences are mostly characterized by occlusions, irregular motion, and illumination changes, proves that our method is robust to those criticalities. OCSORT provides the best results in terms of "IDswitch," but it also provides the highest number of false negatives, meaning that it ignores a large number of instances and so is not suitable for agronomic operations. In the \textit{Overview} sequences our approach consistently exhibits stronger capabilities compared to the other trackers, with a few exceptions where SORT remains competitive in some metrics by a small margin. This happens because SORT excels particularly when the assumption of linear motion holds, as in the \textit{Overview2} sequence, because it consists of a slow and regular walk parallel to the vineyard. However, this assumption is overly restrictive and does not accurately represent real-world robotics scenarios, where robots must move and approach crops closely to perform various operations. The results also show that AgriSORT consistently outperforms all its competitors in terms of MT and ML tracks, underscoring its robust association capabilities. These metrics are important for agronomic tasks like yield estimation, where it is important to not lose tracks, otherwise the same grape bunch gets counted more than once, leading to a wrong estimate of the workload.

Regarding overall results, we expected BoT-SORT and StrongSORT to perform poorly because they use deep models to extract visual information from objects that have not been finetuned on our data. Moreover, as stated in \cite{1_lettuceTrack}, tracking plants presents distinct challenges compared to tracking humans due to the inherent similarities between different instances and the absence of distinct characteristics for disambiguation. In contrast, our method exhibits greater resilience by relying solely on easily extractable motion information within a static environment. In the comparison with OC-SORT, SORT, and ByteTrack, all of which rely solely on the detector, our association strategy excels within our specific context, primarily due to its robustness in handling abrupt motion changes. Visual representations of the performance disparities among these various trackers can be found in Fig. \ref{fig::comparison}.




\begin{table}[t]
\centering
\footnotesize
\vspace{5pt}
\caption{\footnotesize Comparison of the performance of different techniques to estimate the camera motion.}
\begin{adjustbox}{width=0.45\textwidth}
\begin{tabular}{|c|c|c|c|c|c|}
\hline
Sequence & Technique & MOTA↑ & IDF1↑  & HOTA↑ & FPS↑\\
\hline
CloseUp1 & LK + Aff & 65.93 & 72.00 & 48.71 & \textbf{25.93}\\ 
    & ORB + Aff & 66.59 & 72.83 & 52.55 & 20.92\\
    & LK + hom & 64.85 & 71.53 & 51.90 & 25.80\\ 
    & ORB + hom & \textbf{67.28} & \textbf{76.32} & \textbf{53.80} & 19.71\\ 
\hline
CloseUp2 & LK + Aff & 66.13 & 73.00 & 56.08 & \textbf{66.09}\\ 
    & ORB + Aff & \textbf{66.32} & 73.10 & 56.12 & 25.19\\
    & LK + hom & 64.59 & \textbf{74.99} & \textbf{57.24} & 63.01\\ 
    & ORB + hom & 65.48 & 74.64 & 56.64 & 25.01\\ 
\hline
Overview1 & LK + Aff & \textbf{62.21} & 73.72 & 52.74 & \textbf{55.99} \\ 
    & ORB + Aff & 62.20 & \textbf{74.05} & \textbf{53.14} & 23.93\\
    & LK + hom & 60.19 & 70.55 & 49.85 & 53.36\\ 
    & ORB + hom & 60.48 & 71.03 & 51.54 & 22.79\\ 
\hline
Overview2 & LK + Aff & \textbf{45.08} & \textbf{56.88} & \textbf{41.33} & \textbf{55.43} \\ 
    & ORB + Aff & 41.75 & 54.53 & 38.75 & 25.42\\
    & LK + hom & 45.08 & 54.66 & 40.27 & 50.35\\ 
    & ORB + hom & 39.39 & 57.12 & 40.39 & 24.75\\ 
\hline
\end{tabular}
\end{adjustbox}

\label{tab:ablation}
\end{table}

We also conducted a study to compare different strategies for estimating the camera motion. In particular, we explored the possibility of substituting the affine transform with a homography matrix and the Lucas-Kanade Optical Flow method with the ORB feature matching. The additional investigation results are displayed in Table \ref{tab:ablation} and show that there is no clear trend demonstrating the dominance of a single technique over the others in terms of metrics, and that the results depend on the characteristics of the single sequence. Despite the minimal differences, the affine transform for motion estimation is superior in the "Overview" sequences compared to homography because the motion is mainly linear and parallel to the crops, which corresponds to small scale changes. On the other hand, homography works better in the "CloseUp" sequences, because they are characterized by the toughest motion, including fast lateral motion and scale changes. Since the metrics are similar, the most important parameter to consider is the computational speed in Frames per Second (FPS) indicating the speed the algorithm takes to process the input, which is vital in robotics applications. The time indicated in the table does not include the detection processing time since it is the same for all techniques and is about 10 ms. The results show that the ORB feature extraction method is much slower than the LK method by an average of 34 FPS in almost all sequences except for the \textit{CloseUp1} sequence, where the speed difference shrinks to only 5 FPS due to some singularities. In our implementation we provide the possibility to change the techniques as parameters, but we propose the combination of affine transform and Lucas-Kanade method for feature extraction as our main pipeline since it is the faster method.

\section{Conclusion}
\label{sec::conclusions}
In this paper, we present AgriSORT, a MOT approach for robotics in precision agriculture, light, fast, and flexible enough that it can be applied to different types of crops given a working detection algorithm. We propose a different formulation of the Kalman Filter specific for the agricultural case, where objects of interest are static and the only source of motion is the dynamic camera. We conducted experiments on real data collected and annotated to validate our method. We compared AgriSORT to other SOTA approaches for MOT to demonstrate that, in this context, it performs better while running at high-speed, confirming its real-time deployment. Future directions include improving the method for estimating the camera motion using learning-based techniques and extending the tracker to mixed classes of objects, including moving ones like people or tractors, which are helpful from an operational perspective.

\AtNextBibliography{\footnotesize}
\printbibliography

@ARTICLE{1_lettuceTrack,
  
AUTHOR={Hu, Nan and Su, Daobilige and Wang, Shuo and Nyamsuren, Purevdorj and Qiao, Yongliang and Jiang, Yu and Cai, Yu},   
	 
TITLE={LettuceTrack: Detection and tracking of lettuce for robotic precision spray in agriculture},      
	
JOURNAL={Frontiers in Plant Science},      
	
VOLUME={13},           
	
YEAR={2022},      
	  
URL={https://www.frontiersin.org/articles/10.3389/fpls.2022.1003243},       
	
DOI={10.3389/fpls.2022.1003243},      
	
ISSN={1664-462X}
}

@article{2_farmMot,
title = {Multiple object tracking in farmland based on fusion point cloud data},
journal = {Computers and Electronics in Agriculture},
volume = {200},
pages = {107259},
year = {2022},
issn = {0168-1699},
doi = {https://doi.org/10.1016/j.compag.2022.107259},
url = {https://www.sciencedirect.com/science/article/pii/S0168169922005725},
author = {Yuhan Ji and Cheng Peng and Shichao Li and Bin Chen and Yanlong Miao and Man Zhang and Han Li}
}

@Article{3_s21196657,
AUTHOR = {Osman, Youssef and Dennis, Reed and Elgazzar, Khalid},
TITLE = {Yield Estimation and Visualization Solution for Precision Agriculture},
JOURNAL = {Sensors},
VOLUME = {21},
YEAR = {2021},
NUMBER = {19},
ARTICLE-NUMBER = {6657},
URL = {https://www.mdpi.com/1424-8220/21/19/6657},
PubMedID = {34640977},
ISSN = {1424-8220},
DOI = {10.3390/s21196657}
}

@Article{4_applmech3030049,
AUTHOR = {Botta, Andrea and Cavallone, Paride and Baglieri, Lorenzo and Colucci, Giovanni and Tagliavini, Luigi and Quaglia, Giuseppe},
TITLE = {A Review of Robots, Perception, and Tasks in Precision Agriculture},
JOURNAL = {Applied Mechanics},
VOLUME = {3},
YEAR = {2022},
NUMBER = {3},
PAGES = {830--854},
URL = {https://www.mdpi.com/2673-3161/3/3/49},
ISSN = {2673-3161},
DOI = {10.3390/applmech3030049}
}

@article{5_article,
author = {Halstead, Michael and Ahmadi, Alireza and Smitt, Claus and Schmittmann, Oliver and Mccool, Chris},
year = {2021},
month = {12},
pages = {},
title = {Crop Agnostic Monitoring Driven by Deep Learning},
volume = {12},
journal = {Frontiers in Plant Science},
doi = {10.3389/fpls.2021.786702}
}

@article{6_CIARFUGLIA2023107624,
title = {Weakly and semi-supervised detection, segmentation and tracking of table grapes with limited and noisy data},
journal = {Computers and Electronics in Agriculture},
volume = {205},
pages = {107624},
year = {2023},
issn = {0168-1699},
doi = {https://doi.org/10.1016/j.compag.2023.107624},
url = {https://www.sciencedirect.com/science/article/pii/S0168169923000121},
author = {Thomas A. Ciarfuglia and Ionut M. Motoi and Leonardo Saraceni and Mulham Fawakherji and Alberto Sanfeliu and Daniele Nardi}
}

@inproceedings{Bewley_2016,
	doi = {10.1109/icip.2016.7533003},
  
	url = {https://doi.org/10.11092Ficip.2016.7533003},
  
	year = 2016,
	month = {9},
  
	publisher = {{IEEE}
},
  
	author = {Alex Bewley and Zongyuan Ge and Lionel Ott and Fabio Ramos and Ben Upcroft},
  
	title = {Simple online and realtime tracking},
  
	booktitle = {2016 {IEEE} International Conference on Image Processing ({ICIP})}
}

@inproceedings{wojke2017simple,
  title={Simple online and realtime tracking with a deep association metric},
  author={Wojke, Nicolai and Bewley, Alex and Paulus, Dietrich},
  booktitle={2017 IEEE international conference on image processing (ICIP)},
  pages={3645--3649},
  year={2017},
  organization={IEEE}
}

@misc{wang2020realtime,
      title={Towards Real-Time Multi-Object Tracking}, 
      author={Zhongdao Wang and Liang Zheng and Yixuan Liu and Yali Li and Shengjin Wang},
      year={2020},
      eprint={1909.12605},
      archivePrefix={arXiv},
      primaryClass={cs.CV}
}

@INPROCEEDINGS{Geiger2012CVPR,
  author = {Andreas Geiger and Philip Lenz and Raquel Urtasun},
  title = {Are we ready for Autonomous Driving? The KITTI Vision Benchmark Suite},
  booktitle = {Conference on Computer Vision and Pattern Recognition (CVPR)},
  year = {2012}
}

@article{MOTChallenge2015,
	title = {{MOTC}hallenge 2015: {T}owards a Benchmark for Multi-Target Tracking},
	shorttitle = {MOTChallenge 2015},
	url = {http://arxiv.org/abs/1504.01942},
	journal = {arXiv:1504.01942 [cs]},
	author = {Leal-Taix\'{e}, L. and Milan, A. and Reid, I. and Roth, S. and Schindler, K.},
	month = apr,
	year = {2015},
	note = {arXiv: 1504.01942}
}

@article{MOTChallenge20,
    title={MOT20: A benchmark for multi object tracking in crowded scenes},
    shorttitle = {MOT20},
	url = {http://arxiv.org/abs/1906.04567},
	journal = {arXiv:2003.09003[cs]},
	author = {Dendorfer, P. and Rezatofighi, H. and Milan, A. and Shi, J. and Cremers, D. and Reid, I. and Roth, S. and Schindler, K. and Leal-Taix\'{e}, L. },
	month = mar,
	year = {2020},
	note = {arXiv: 2003.09003}
}

@article{liu2023orb,
  title={ORB-Livox: A real-time dynamic system for fruit detection and localization},
  author={Liu, Tianhao and Kang, Hanwen and Chen, Chao},
  journal={Computers and Electronics in Agriculture},
  volume={209},
  pages={107834},
  year={2023},
  publisher={Elsevier}
}

@article{underwood2015lidar,
  title={Lidar-based tree recognition and platform localization in orchards},
  author={Underwood, James P and Jagbrant, Gustav and Nieto, Juan I and Sukkarieh, Salah},
  journal={Journal of Field Robotics},
  volume={32},
  number={8},
  pages={1056--1074},
  year={2015},
  publisher={Wiley Online Library}
}

@inproceedings{bargoti2017deep,
  title={Deep fruit detection in orchards},
  author={Bargoti, Suchet and Underwood, James},
  booktitle={2017 IEEE international conference on robotics and automation (ICRA)},
  pages={3626--3633},
  year={2017},
  organization={IEEE}
}

@article{mu2020intact,
  title={Intact detection of highly occluded immature tomatoes on plants using deep learning techniques},
  author={Mu, Yue and Chen, Tai-Shen and Ninomiya, Seishi and Guo, Wei},
  journal={Sensors},
  volume={20},
  number={10},
  pages={2984},
  year={2020},
  publisher={MDPI}
}

@article{mai2020faster,
  title={Faster R-CNN with classifier fusion for automatic detection of small fruits},
  author={Mai, Xiaochun and Zhang, Hong and Jia, Xiao and Meng, Max Q-H},
  journal={IEEE Transactions on Automation Science and Engineering},
  volume={17},
  number={3},
  pages={1555--1569},
  year={2020},
  publisher={IEEE}
}

@article{liu2020yolo,
  title={YOLO-tomato: A robust algorithm for tomato detection based on YOLOv3},
  author={Liu, Guoxu and Nouaze, Joseph Christian and Touko Mbouembe, Philippe Lyonel and Kim, Jae Ho},
  journal={Sensors},
  volume={20},
  number={7},
  pages={2145},
  year={2020},
  publisher={MDPI}
}

@article{du2023strongsort,
  title={Strongsort: Make deepsort great again},
  author={Du, Yunhao and Zhao, Zhicheng and Song, Yang and Zhao, Yanyun and Su, Fei and Gong, Tao and Meng, Hongying},
  journal={IEEE Transactions on Multimedia},
  year={2023},
  publisher={IEEE}
}

@inproceedings{zhang2022bytetrack,
  title={Bytetrack: Multi-object tracking by associating every detection box},
  author={Zhang, Yifu and Sun, Peize and Jiang, Yi and Yu, Dongdong and Weng, Fucheng and Yuan, Zehuan and Luo, Ping and Liu, Wenyu and Wang, Xinggang},
  booktitle={European Conference on Computer Vision},
  pages={1--21},
  year={2022},
  organization={Springer}
}

@inproceedings{du2021giaotracker,
  title={Giaotracker: A comprehensive framework for mcmot with global information and optimizing strategies in visdrone 2021},
  author={Du, Yunhao and Wan, Junfeng and Zhao, Yanyun and Zhang, Binyu and Tong, Zhihang and Dong, Junhao},
  booktitle={Proceedings of the IEEE/CVF International conference on computer vision},
  pages={2809--2819},
  year={2021}
}

@misc{cao2023observationcentric,
      title={Observation-Centric SORT: Rethinking SORT for Robust Multi-Object Tracking}, 
      author={Jinkun Cao and Jiangmiao Pang and Xinshuo Weng and Rawal Khirodkar and Kris Kitani},
      year={2023},
      eprint={2203.14360},
      archivePrefix={arXiv},
      primaryClass={cs.CV}
}

@article{zhang2021fairmot,
  title={Fairmot: On the fairness of detection and re-identification in multiple object tracking},
  author={Zhang, Yifu and Wang, Chunyu and Wang, Xinggang and Zeng, Wenjun and Liu, Wenyu},
  journal={International Journal of Computer Vision},
  volume={129},
  pages={3069--3087},
  year={2021},
  publisher={Springer}
}

@inproceedings{bergmann2019tracking,
  title={Tracking without bells and whistles},
  author={Bergmann, Philipp and Meinhardt, Tim and Leal-Taixe, Laura},
  booktitle={Proceedings of the IEEE/CVF International Conference on Computer Vision},
  pages={941--951},
  year={2019}
}

@article{zhou2020tracking,
  title={Tracking Objects as Points},
  author={Zhou, Xingyi and Koltun, Vladlen and Kr{\"a}henb{\"u}hl, Philipp},
  journal={ECCV},
  year={2020}
}

@inproceedings{tokmakov2021learning,
  title={Learning to Track with Object Permanence},
  author={Tokmakov, Pavel and Li, Jie and Burgard, Wolfram and Gaidon, Adrien},
  booktitle={ICCV},
  year={2021}
}

@article{sun2020transtrack,
  title={Transtrack: Multiple object tracking with transformer},
  author={Sun, Peize and Cao, Jinkun and Jiang, Yi and Zhang, Rufeng and Xie, Enze and Yuan, Zehuan and Wang, Changhu and Luo, Ping},
  journal={arXiv preprint arXiv:2012.15460},
  year={2020}
}

@inproceedings{meinhardt2022trackformer,
  title={Trackformer: Multi-object tracking with transformers},
  author={Meinhardt, Tim and Kirillov, Alexander and Leal-Taixe, Laura and Feichtenhofer, Christoph},
  booktitle={Proceedings of the IEEE/CVF conference on computer vision and pattern recognition},
  pages={8844--8854},
  year={2022}
}

@article{li2022simpletrack,
  title={Simpletrack: Rethinking and improving the jde approach for multi-object tracking},
  author={Li, Jiaxin and Ding, Yan and Wei, Hua-Liang and Zhang, Yutong and Lin, Wenxiang},
  journal={Sensors},
  volume={22},
  number={15},
  pages={5863},
  year={2022},
  publisher={MDPI}
}

@inproceedings{lu2020retinatrack,
  title={Retinatrack: Online single stage joint detection and tracking},
  author={Lu, Zhichao and Rathod, Vivek and Votel, Ronny and Huang, Jonathan},
  booktitle={Proceedings of the IEEE/CVF conference on computer vision and pattern recognition},
  pages={14668--14678},
  year={2020}
}

@article{aharon2022bot,
  title={BoT-SORT: Robust associations multi-pedestrian tracking},
  author={Aharon, Nir and Orfaig, Roy and Bobrovsky, Ben-Zion},
  journal={arXiv preprint arXiv:2206.14651},
  year={2022}
}

@software{CVAT_ai_Corporation_Computer_Vision_Annotation_2022,
author = {{CVAT.ai Corporation}},
license = {MIT},
month = sep,
title = {{Computer Vision Annotation Tool (CVAT)}},
url = {https://github.com/opencv/cvat},
version = {2.2.0},
year = {2022}
}

@inproceedings{lucas1981iterative,
  title={An iterative image registration technique with an application to stereo vision},
  author={Lucas, Bruce D and Kanade, Takeo},
  booktitle={IJCAI'81: 7th international joint conference on Artificial intelligence},
  volume={2},
  pages={674--679},
  year={1981}
}

@article{zhaoxin2022design,
  title={Design a robot system for tomato picking based on yolo v5},
  author={Zhaoxin, Guan and Han, Li and Zhijiang, Zuo and Libo, Pan},
  journal={IFAC-PapersOnLine},
  volume={55},
  number={3},
  pages={166--171},
  year={2022},
  publisher={Elsevier}
}

@article{milan2016mot16,
  title={MOT16: A benchmark for multi-object tracking},
  author={Milan, Anton and Leal-Taix{\'e}, Laura and Reid, Ian and Roth, Stefan and Schindler, Konrad},
  journal={arXiv preprint arXiv:1603.00831},
  year={2016}
}

@article{luiten2021hota,
  title={Hota: A higher order metric for evaluating multi-object tracking},
  author={Luiten, Jonathon and Osep, Aljosa and Dendorfer, Patrick and Torr, Philip and Geiger, Andreas and Leal-Taix{\'e}, Laura and Leibe, Bastian},
  journal={International journal of computer vision},
  volume={129},
  pages={548--578},
  year={2021},
  publisher={Springer}
}

@article{bernardin2008evaluating,
  title={Evaluating multiple object tracking performance: the clear mot metrics},
  author={Bernardin, Keni and Stiefelhagen, Rainer},
  journal={EURASIP Journal on Image and Video Processing},
  volume={2008},
  pages={1--10},
  year={2008},
  publisher={Springer}
}

@article{ciarfuglia2023weakly,
  title={Weakly and semi-supervised detection, segmentation and tracking of table grapes with limited and noisy data},
  author={Ciarfuglia, Thomas A and Motoi, Ionut M and Saraceni, Leonardo and Fawakherji, Mulham and Sanfeliu, Alberto and Nardi, Daniele},
  journal={Computers and Electronics in Agriculture},
  volume={205},
  pages={107624},
  year={2023},
  publisher={Elsevier}
}

@inproceedings{ciarfuglia2022pseudo,
  title={Pseudo-label generation for agricultural robotics applications},
  author={Ciarfuglia, Thomas A and Motoi, Ionut Marian and Saraceni, Leonardo and Nardi, Daniele},
  booktitle={Proceedings of the IEEE/CVF Conference on Computer Vision and Pattern Recognition},
  pages={1686--1694},
  year={2022}
}

@misc{luiten2020trackeval,
  author =       {Jonathon Luiten, Arne Hoffhues},
  title =        {TrackEval},
  howpublished = {\url{https://github.com/JonathonLuiten/TrackEval}},
  year =         {2020}
}

@inproceedings{haug2014plant,
  title={Plant classification system for crop/weed discrimination without segmentation},
  author={Haug, Sebastian and Michaels, Andreas and Biber, Peter and Ostermann, J{\"o}rn},
  booktitle={IEEE winter conference on applications of computer vision},
  pages={1142--1149},
  year={2014},
  organization={IEEE}
}

@inproceedings{lottes2017uav,
  title={UAV-based crop and weed classification for smart farming},
  author={Lottes, Philipp and Khanna, Raghav and Pfeifer, Johannes and Siegwart, Roland and Stachniss, Cyrill},
  booktitle={2017 IEEE international conference on robotics and automation (ICRA)},
  pages={3024--3031},
  year={2017},
  organization={IEEE}
}

@article{milioto2017real,
  title={Real-time blob-wise sugar beets vs weeds classification for monitoring fields using convolutional neural networks},
  author={Milioto, Andres and Lottes, Philipp and Stachniss, Cyrill},
  journal={ISPRS Annals of the Photogrammetry, Remote Sensing and Spatial Information Sciences},
  volume={4},
  pages={41--48},
  year={2017},
  publisher={Copernicus GmbH}
}

@article{saleem2021automation,
  title={Automation in agriculture by machine and deep learning techniques: A review of recent developments},
  author={Saleem, Muhammad Hammad and Potgieter, Johan and Arif, Khalid Mahmood},
  journal={Precision Agriculture},
  volume={22},
  pages={2053--2091},
  year={2021},
  publisher={Springer}
}

@article{jin2022novel,
  title={A novel deep learning-based method for detection of weeds in vegetables},
  author={Jin, Xiaojun and Sun, Yanxia and Che, Jun and Bagavathiannan, Muthukumar and Yu, Jialin and Chen, Yong},
  journal={Pest Management Science},
  volume={78},
  number={5},
  pages={1861--1869},
  year={2022},
  publisher={Wiley Online Library}
}

@article{kuhn1955hungarian,
  title={The Hungarian method for the assignment problem},
  author={Kuhn, Harold W},
  journal={Naval research logistics quarterly},
  volume={2},
  number={1-2},
  pages={83--97},
  year={1955},
  publisher={Wiley Online Library}
}

\end{document}